\documentclass{article}

\usepackage[preprint]{neurips_2026}

\usepackage[utf8]{inputenc} 
\usepackage[T1]{fontenc}    
\usepackage{hyperref}       
\usepackage{url}            
\usepackage{booktabs}       
\usepackage{amsfonts}       
\usepackage{nicefrac}       
\usepackage{microtype}      
\usepackage{xcolor}         

\usepackage{amsmath}    
\usepackage{amssymb}     
\usepackage{mathtools}   
\usepackage{bm}
\usepackage{enumitem}
\usepackage{multirow}
\usepackage{array}
\usepackage{subcaption}
\usepackage{wrapfig}
\usepackage[table]{xcolor}

\title{LoopVLA: Learning Sufficiency in Recurrent Refinement for Vision-Language-Action Models}

\author{
Boyang Shen$^{1}$, Kaixiang Yang$^{1}$, Hao Wang$^{1}$, Qiuyu Yu$^{2}$, Qiang Xie$^{2}$, Qiang Li$^{1,*}$, Zhiwei Wang$^{1,*}$ \\
$^{1}$Huazhong University of Science and Technology \\
$^{2}$Wuhan United Imaging Surgical Co.,Ltd. (UIS)\\
$^{*}$Corresponding authors:\texttt{\{liqiang8,zwwang\}@hust.edu.cn}
}

\begin{document}

\maketitle

\begin{abstract}
Current Vision-Language-Action (VLA) models typically treat the deepest representation of a vision-language backbone as universally optimal for action prediction. However, robotic manipulation is composed of many frequent closed-loop spatial adjustments, for which excessive abstraction may waste computation and weaken low-level geometric cues essential for precise control. Existing early-exit strategies attempt to reduce computation by stopping at predefined layers or applying heuristic rules such as action consistency, but they do not directly answer when a representation is actually sufficient for action.
In this paper, we present \textbf{LoopVLA}, a recurrent VLA architecture that jointly learns representation refinement, action prediction, and sufficiency estimation.
LoopVLA iteratively applies a shared Transformer block to refine multimodal tokens, and at each iteration produces both a candidate action and a sufficiency score that estimates whether further refinement is necessary. By sharing parameters across iterations, LoopVLA decouples refinement from absolute layer indices and grounds sufficiency estimation in the evolving representation itself.
Since sufficiency has no direct supervision, we introduce a self-supervised distribution alignment objective, where intermediate confidence scores are trained to match the relative action quality across refinement steps, thereby linking sufficiency learning to policy optimization signals.
Experiments on LIBERO, LIBERO-Plus, and VLA-Arena show that LoopVLA pushes the efficiency-performance frontier of VLA policies, reducing parameters by 45\% and improving inference throughput by up to 1.7 times while matching or outperforming strong baselines in task success.

\end{abstract}

\section{Introduction}
Vision-Language-Action (VLA) models represent a significant step toward generalist robotic manipulation, using large-scale multimodal pretraining to bridge perceptual understanding and physical interaction~\cite{brohan2023rt1,zitkovich2023rt,kim2025openvla,mees2024octo}.
A typical VLA system builds upon a vision-language model (VLM) to encode visual observations and language instructions into latent representations, and then employs an action head that maps these representations to continuous control outputs~\cite{radford2021learning,chen2022pali,alayrac2022flamingo,li2023blip,liu2023visual}.
Most existing designs follow a \emph{late-output} paradigm, where only the final-layer representation of the VLM is used for action prediction.

Despite its simplicity, the late-output paradigm implicitly assumes that the deepest, most abstract representation is universally suitable for all action decisions.
This assumption, however, is not well aligned with the heterogeneous nature of embodied decision‑making~\cite{kurtzer2008long,pruszynski2012optimal}.
In robotic manipulation, actions vary substantially from simple, reactive adjustments to complex, fine‑grained manipulations.
The latter may demand richer semantic abstraction, whereas the former depends more directly on low-level spatial, geometric, and reactive cues.

This observation motivates a key problem in VLA models: \emph{representational sufficiency}.
For action prediction, the objective is not necessarily to reach the deepest representation, but to obtain a representation that is \emph{sufficient} for the current control decision.
An under-refined representation may lead to inaccurate actions, while an over-processed representation may dilute task-relevant cues and reduce inference efficiency.
This perspective also aligns with biological motor control, where simple reflexive responses rely on short neural pathways, while complex behaviors engage deeper processing circuits~\cite{kurtzer2008long,pruszynski2012optimal}.

\begin{figure}
  \centering
  \includegraphics[width=\columnwidth]{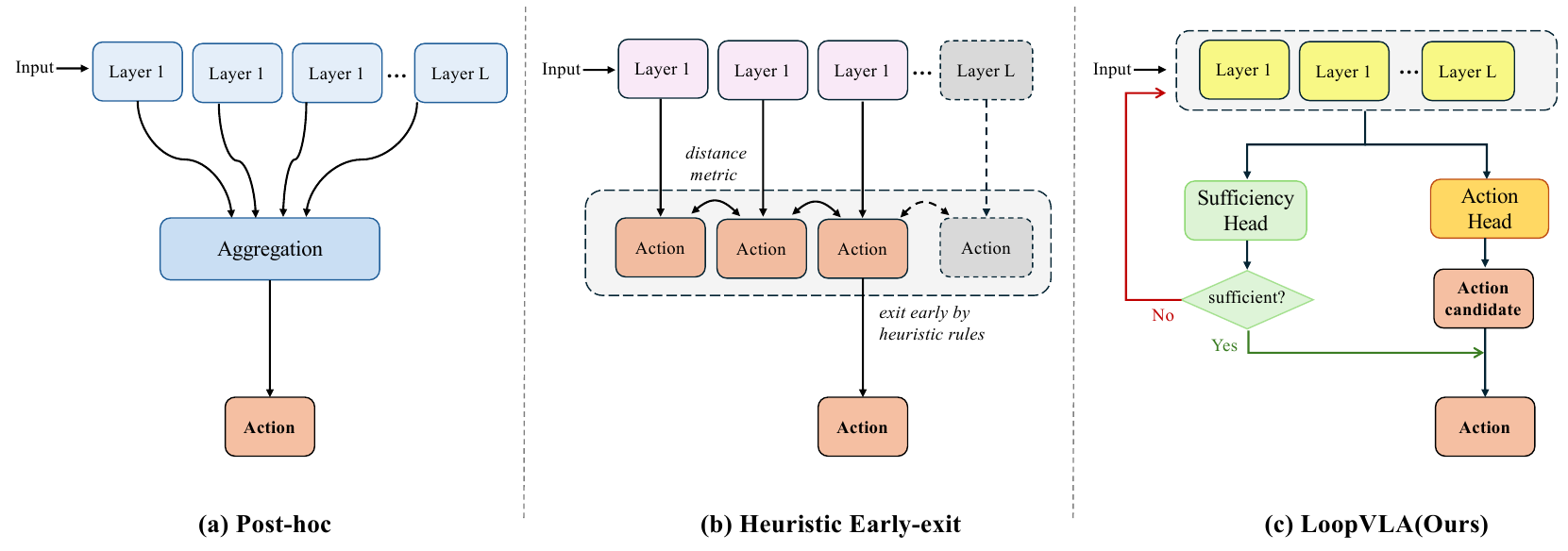}
  \caption{Comparison of intermediate-representation paradigms in VLA models.
(a) Post-hoc methods aggregate or select intermediate features after full-depth computation.
(b) Heuristic early-stop methods terminate computation using predefined criteria.
(c) LoopVLA jointly models iterative refinement and sufficiency estimation through a shared recurrent Transformer.}
  \label{fig:compare}
\end{figure}

Recent efforts have begun to exploit intermediate representations to overcome the rigidity of the late-output paradigm.
These methods can be broadly organized into two categories.
The first category requires a full forward pass through the VLM and then performs \emph{post-hoc aggregation or selection} of features from multiple layers via auxiliary mechanisms~\cite{shukor2025smolvla,Black2025AV}, as illustrated in Fig.~\ref{fig:compare}(a).
Although such methods can harness multi-level information beyond the final layer, they treat representation selection as a decoupled, retrospective step and invariably incur the full computational cost of processing all layers.

The second category relies on \emph{early-exit} strategies that stop computation before the final layer.
Static approaches~\cite{reuss2025flower,nvidia2025gr00tn1} directly use a predefined intermediate layer, remaining inherently rigid to varying task demands.
Dynamic approaches, by contrast, allow the model to exit at different depths depending on the input.
However, they typically rely on heuristic rules such as action consistency~\cite{yue2024deer}, which are not explicitly aligned with action quality and may not reliably indicate representational sufficiency.


In this work, we argue that representational sufficiency should be treated as an \emph{intrinsic property} of the policy itself. This calls for a unified formulation in which representation refinement and action prediction are jointly modeled, enabling the policy to dynamically determine how much computation is needed.

To this end,we propose \textbf{LoopVLA}, a VLA architecture that models representation refinement and sufficiency \emph{in situ} through a recurrent formulation.
Specifically, we introduce a dual-head design: at each processing stage, an action head produces a candidate control output, while a gating head estimates a confidence score that quantifies the sufficiency of the current representation for the action at hand. This allows sufficiency to be explicitly modeled and directly tied to decision quality.

A straightforward implementation might attach such heads to each layer of the current VLM. However, this risks introducing biases tied to absolute layer indices, rather than grounding sufficiency in representational content, as illustrated in Fig.~\ref{fig:compare}(b).
To avoid this, we adopt a recurrent parameter-sharing design: a shallow Transformer model is applied iteratively, and the two heads are attached to each iteration, as illustrated in Fig.~\ref{fig:compare}(c).. This loop structure decouples refinement from absolute depth, compelling the model to base sufficiency estimates on the evolving features themselves.

As no ground-truth labels exist for representational sufficiency, we learn it through distribution alignment.
During training, a fixed number of iterations is performed, and both the negative action losses and the confidence scores are normalized across iterations to form probability distributions. The confidence distribution is then trained to match the loss-derived distribution via a cross-entropy objective. This formulation treats the loss profile as soft supervision, enabling the model to recognize sufficient representations without any external annotation.

The main contributions of this work are summarized as follows:

\begin{enumerate}[leftmargin=*, noitemsep, topsep=0pt]

\item Building upon prior representational sufficiency studies, we reformulate VLA representation learning as an iterative refinement process with a shared recurrent Transformer, and introduce a distribution alignment objective for learning sufficiency estimation from action optimization signals.

\item We propose LoopVLA, a recurrent VLA architecture that replaces the conventional fixed-depth Transformer with a shared looped Transformer for progressive refinement. The proposed design reduces model size by \textbf{45\%} while maintaining strong performance.

\item We introduce a dual-head design with an action head and a sufficiency head, enabling sufficiency estimation directly from evolving representations rather than heuristic early-exit criteria.

\item We evaluate LoopVLA on LIBERO, LIBERO-Plus, and VLA-Arena, demonstrating improved inference efficiency while maintaining or improving policy performance over strong baselines, including up to \textbf{1.7$\times$} higher inference throughput in extreme settings with substantially fewer parameters.

\end{enumerate}

\section{Related Work}


\paragraph{Vision-Language-Action~Models.}
\begin{wrapfigure}{r}{5.7cm}
  \vspace{-0.45cm}
  \centering
  \includegraphics[width=0.9\linewidth]{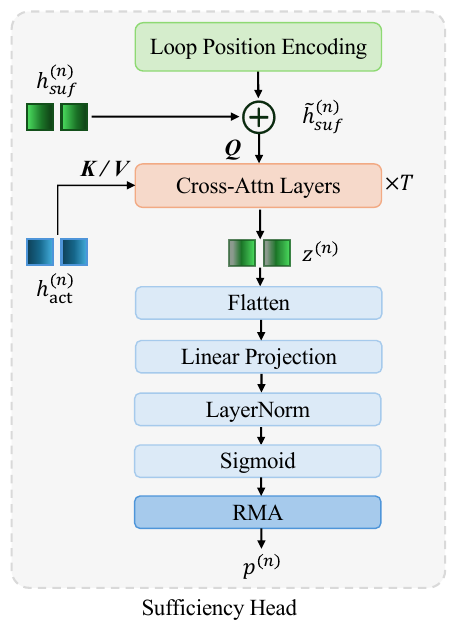}
  \caption{Sufficiency Head architecture. At iteration $n$, sufficiency tokens attend to action tokens through cross-attention layers with loop positional encoding to predict the halting probability $p^{(n)}$. RMA denotes the Remaining Mass Allocation mechanism.}
  \label{fig:detail}
  \vspace{-1cm}
\end{wrapfigure}
Vision-Language-Action (VLA) models have emerged as a promising paradigm for generalist robot manipulation by leveraging large-scale multimodal data and pretrained vision-language models (VLMs) \cite{karamcheti2024prismatic, steiner2024paligemma, bai2025qwen3}. A typical VLA system encodes visual observations and language instructions into latent representations using a VLM backbone, followed by a policy head that maps these representations to control actions \cite{kawaharazuka2025vision}.

Recent efforts have focused on scaling data, model capacity, and task diversity, leading to improved generalization across tasks and environments \cite{cheang2024gr, bu2025univla}. In most existing approaches, action prediction is primarily based on the final-layer representations of the backbone, implicitly assuming that deeper representations are more suitable for decision making. While effective in practice, this design largely treats the output of the backbone as a fixed representation for downstream control.

\paragraph{Layer-wise Representation Readout.}
Prior work has shown that representations at different layers of VLMs encode information at varying levels of abstraction, from low-level perceptual cues to high-level semantic concepts \cite{chen2025rethinking}. To better exploit these intermediate representations, several studies explore layer-wise readout strategies for action prediction. One line of work attaches the action head to a fixed layer, assuming a single depth provides optimal features across inputs \cite{reuss2025flower,nvidia2025gr00tn1}. While efficient, this design is inherently input-agnostic and may be suboptimal under varying task complexities.

Alternatively, multi-layer aggregation methods combine representations from different depths via attention or learned fusion modules \cite{Black2025AV, shukor2025smolvla, chen2025internvla}. Although more expressive, these approaches introduce additional computational overhead and training complexity. Despite their differences, both lines of work treat depth as a static architectural choice, overlooking the role of progressive refinement in shaping representations for accurate action prediction.

\paragraph{Iterative Refinement Transformer.}
Iterative formulations of Transformer architectures have been explored through recurrent designs, where a shared block is applied multiple times to progressively refine representations. Prior works, such as Adaptive Computation Time and Universal Transformer, regulate or expose the number of refinement steps based on intermediate signals \cite{graves2016adaptive, dehghani2019universal}, offering a view of depth as an unrolled iterative process rather than as a fixed architectural parameter.

More recently, this paradigm has been extended to vision-language and vision-language-action models. Several works investigate recurrent computation or iterative refinement within pretrained multimodal Transformers to improve representation quality and enable additional test-time computation \cite{saunshi2025reasoning, zhu2025scaling}. RD-VLA \cite{tur2026recurrent} further explores this idea for robotic manipulation by repeatedly applying a weight-shared recurrent module to an action head, performing iterative refinement in latent space before decoding the final action. It also leverages the convergence of intermediate actions to adapt the number of recurrent steps at inference time, enabling input-dependent test-time computation.

While RD-VLA demonstrates the potential of recurrent computation for action prediction, its recurrent computation is confined to the action head. In contrast, LoopVLA introduces recurrence directly into the pretrained VLM decoder, repeatedly reusing a subset of Transformer layers to progressively refine the multimodal representation before action prediction. This design treats the depth of the VLM itself as a dynamically adjustable computational resource, rather than repeatedly refining only the downstream action representation. Moreover, LoopVLA explicitly learns a representation sufficiency estimator to determine whether the current multimodal representation is sufficient for action prediction, enabling adaptive computation based on representation sufficiency rather than action convergence.

\begin{figure}
  \centering
  \includegraphics[width=\columnwidth]{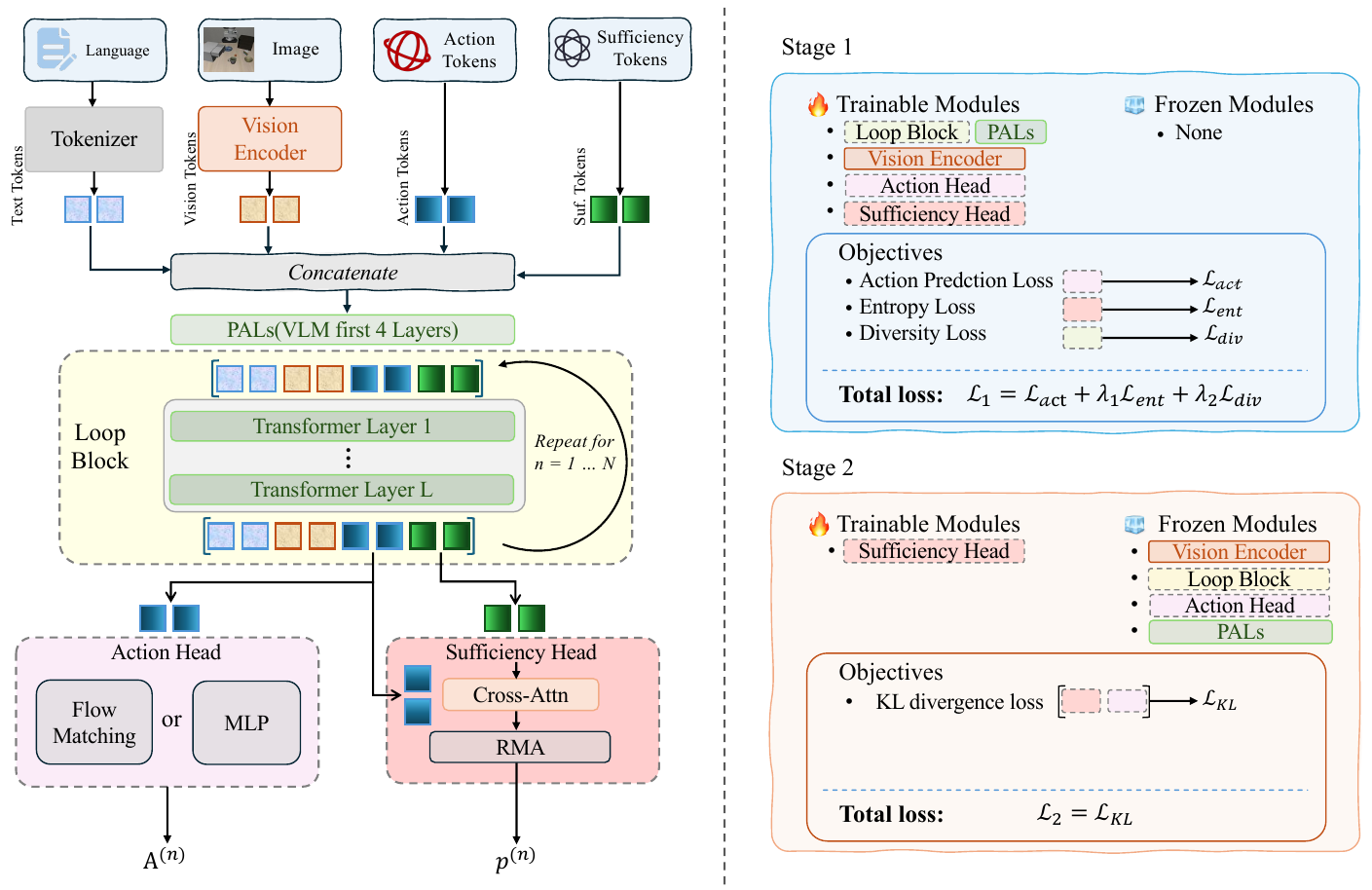}
  \caption{Overview of LoopVLA. Visual, language, action, and sufficiency tokens are processed by perceptual anchor layers followed by a shared recurrent Loop Block for iterative refinement. The action head predicts candidate actions at each iteration, while the sufficiency head estimates halting probabilities through cross-attention and remaining-mass allocation. Training is performed in two stages: joint refinement learning and sufficiency calibration.}
  \label{fig:overview}
\end{figure}

\section{Method}

\subsection{Problem Formulation}
We consider a manipulation dataset with multiple tasks
$\mathcal{D} = \{(\ell_i, \{(o_t, a_t)\}_{t=0}^{T_i})\}_{i=1}^{N}$. 
Each task $i$ is specified by a language instruction $\ell_i$ and a sequence of demonstrations of length $T_i$. 
At each time step $t$, the observation $o_t \in \mathbb{R}^{V \times H \times W \times 3}$ comprises RGB images captured from $V$ camera views, while the action $a \in \mathbb{R}^{J}$ denotes the corresponding control signal, where $J$ is the number of degrees of freedom (DoF).

A typical VLA policy $\pi_\theta$ operates under a Markovian assumption, generating a sequence of future actions conditioned on the current observation and language instruction. 
The inputs are encoded by preserved vision--language model (VLM) layers into a latent representation,
\begin{equation}
   h_t = \mathrm{VLM}(o_t, \ell), 
\end{equation}
which is subsequently mapped to an action sequence by an action head,
\begin{equation}
A_{t:t+c} = \pi_\theta(h_t),
\end{equation}
where $A_{t:t+c} = (a_{t+1}, \ldots, a_{t+c}) \in \mathbb{R}^{c \times J}$ denotes $c$ future actions.
Instead of relying on a fixed latent representation, we formulate representation learning as a dynamic selection process as we proposed. 

\subsection{Model Architecture}

We consider the LoopVLA setting, as illustrated in Fig.~\ref{fig:overview}. We focus on a single-step formulation, where the policy operates on the current observation without explicit temporal dependencies. The policy receives visual observations, which are encoded into token representations via a vision encoder: $h_{\text{vis}} = \mathcal{E}_{\text{vis}}(o)$. Similarly, the language instruction is mapped into text tokens: $h_{\text{txt}} = \mathcal{E}_{\text{txt}}(\ell)$. We further introduce learnable action tokens $h_{\text{act}} \in \mathbb{R}^{N_{\text{act}} \times d}$ and sufficiency tokens $h_{\text{suf}} \in \mathbb{R}^{N_{\text{suf}} \times d}$, which are used to extract action features and loop-dependent representations for sufficiency estimation, respectively.

The tokens are concatenated along the sequence dimension:
$h = [h_{\text{txt}}, h_{\text{vis}}, h_{\text{act}}, h_{\text{suf}}].$
To preserve the early-stage representations of the pretrained LLM, we retain the first four Transformer layers as Perceptual Anchor Layers (PALs), keeping their original architecture unchanged during iterative refinement. The resulting representation is computed as:
$h^{(0)} = \mathrm{PALs}(h),$
where the superscript 0 means the initial state after PALs.

\paragraph{Loop Block.}
We employ a shallow Transformer block as a recurrent module to iteratively refine representations. 
At iteration $n$, the hidden state is updated as
\begin{equation}
h^{(n)} = \mathrm{LoopBlock}\big(h^{(n-1)}; M\big),
\end{equation}
where $\mathrm{LoopBlock}$ denotes a stack of causal Transformer layers with shared parameters, and $M$ is a structured causal attention mask that decouples action and sufficiency tokens. 

Concretely, it follows standard causal masking, with sufficiency tokens fully visible to each other, while enforcing asymmetric visibility: action tokens attend to visual and language tokens, whereas sufficiency tokens attend to all tokens to aggregate global information.

\paragraph{Sufficiency Head.}
As illustrated in Fig.~\ref{fig:detail}, we construct the sufficiency head from the refined representation at iteration $n$. 
We extract the sufficiency and action tokens, denoted as $h_{\text{suf}}^{(n)}$ and $h_{\text{act}}^{(n)}$, respectively. 
A loop-index positional encoding is added to the sufficiency tokens:
\begin{equation}
\tilde{h}_{\text{suf}}^{(n)} = h_{\text{suf}}^{(n)} + PE(n),
\end{equation}
where $PE(n)$ denotes a sinusoidal positional encoding indexed by the iteration step, providing a loop-aware prior over refinement depth. It helps distinguish refinement stages and prevents collapse to trivial identity mappings.

The tokens are then linearly projected to form queries, keys, and values:
$Q\!=\!W_q \tilde{h}_{\text{suf}}^{(n)},
K\!=\!W_k h_{\text{act}}^{(n)}, 
V\!=\!W_v h_{\text{act}}^{(n)}.$
We apply a stack of cross-attention layers:
$z^{(n)} = \mathrm{CrossAttn}^{(T)}(Q, K, V),$
where $\mathrm{CrossAttn}^{(T)}$ denotes $T$ stacked cross-attention layers with shared parameters across iterations. Finally, an MLP followed by a sigmoid produces an intermediate halting score:
\begin{equation}
s^{(n)} = \sigma\!\big(\mathrm{MLP}(z^{(n)})\big),
\end{equation}
where the sigmoid function constrains the output to the range $(0,1)$, enabling a probabilistic interpretation of the halting score.

\paragraph{Remaining Mass Allocation (RMA).}
To ensure proper normalization of sufficiency probabilities across loop iterations, we introduce a remaining mass allocation mechanism. Remaining mass is used to convert intermediate halting scores into a valid distribution over refinement steps. 
At each iteration, the halting probability is computed as
\begin{equation}
p^{(n)} = s^{(n)} \cdot r^{(n)},
\end{equation}
where $r^{(n)}$ denotes the remaining probability mass with an initial value $r^{(1)}=1$, updated recursively by
\begin{equation}
    r^{(n+1)} = r^{(n)} \cdot (1 - s^{(n)}).
\end{equation}
This formulation ensures $\sum_{n=1}^{N} p^{(n)} \le 1$ and enables adaptive halting across iterations.

\paragraph{Action Head.}
At iteration $n$, the action head maps the refined representation to an action chunk:
\begin{equation}
    A^{(n)} = \pi_{\theta}\!\big(h^{(n)}\big),
\end{equation}
where $A^{(n)} \in \mathbb{R}^{c \times J}$ denotes the sequence of $c$ future actions as mentioned. 
The policy head $\pi_{\theta}$ can be instantiated with different parameterizations, such as an OFT-based head or a flow-matching head. 
Each iteration produces a candidate action, whose quality is assessed by the sufficiency head.

\subsection{Dual-Stage Training}
\label{sec:train_detail}

We adopt a dual-stage training strategy to learn both action prediction and sufficiency-aware halting behavior. The first stage jointly optimizes all modules to ensure strong action prediction performance while enabling the sufficiency head to learn a soft halting distribution.

\paragraph{Stage 1: Joint refinement learning.}
We jointly supervise all intermediate predictions to ensure each refinement step produces a meaningful action estimate:
\begin{equation}
\mathcal{L}_{\text{action}} = \sum_{n=1}^{N} \ell(A^{(n)}, \hat{A}),
\end{equation}
where $\hat{A}$ denotes the ground-truth action and $\ell(\cdot,\cdot)$ denotes the $\ell_1$ loss. 
This joint supervision stabilizes training and encourages progressively refined representations.

To further stabilize the halting behavior, we introduce two regularization terms during early training: an entropy regularization to prevent premature collapse, and a diversity regularization to discourage trivial identity mappings:
\begin{equation}
\mathcal{L}_{\text{ent}} = \sum_{n=1}^{N} p^{(n)} \log p^{(n)}, 
\quad
\mathcal{L}_{\text{div}} = \sum_{i<j} \max\big(0, \cos(z^{(i)}, z^{(j)})\big).
\end{equation}
The overall objective is
\begin{equation}
\mathcal{L}_{\text{stage1}} = \mathcal{L}_{\text{action}} + \lambda_1 \mathcal{L}_{\text{ent}} + \lambda_2 \mathcal{L}_{\text{div}}.
\end{equation}

\paragraph{Stage 2: Sufficiency calibration.}
We freeze all modules except the sufficiency head and calibrate the halting distribution. 
This stage aims to decouple action learning from halting prediction, avoiding interference between the two objectives. 
It further sharpens the halting distribution by aligning it with the relative quality of intermediate predictions.
A target distribution over refinement steps is defined as
\begin{equation}
q^{(n)} = \mathrm{softmax}\!\left(-\ell(A^{(n)}, \hat{A}) / \tau\right),
\end{equation}
where $\tau > 0$ controls the sharpness. 
We align the predicted distribution $p^{(n)}$ with $q^{(n)}$ via
\begin{equation}
\mathcal{L}_{\text{stage2}} = \mathrm{KL}(q \,\|\, p),
\end{equation}
where $p = (p^{(1)}, \dots, p^{(N)})$.

\subsection{Inference Recipe}
\label{sec:infer_detail}

At inference time, we utilize the learned halting distribution to select the final action, enabling a flexible trade-off between performance and efficiency.

\textbf{Optimal selection.}
We run all $N$ refinement steps and select the action with the highest probability as the optimal performance strategy :
\begin{equation}
n^* = \arg\max_{n} p^{(n)}, \quad
A = A^{(n^*)}.
\end{equation}

\textbf{Adaptive trade-off.}
To reduce computation, we adopt an early stopping strategy based on the accumulated halting probability. At step $n$, refinement is terminated if either
\begin{equation}
\sum_{k=1}^n p^{(k)} \ge \theta 
\quad \text{or} \quad 
\max_{k \le n} p^{(k)} \ge r^{(n+1)},
\end{equation}
where $\theta$ is a predefined threshold determined heuristically based on the $w\sigma$ rule of a normal distribution (e.g., $\theta \approx 0.68, 0.95, 0.997$ for $w=1,2,3$, respectively).The final action is selected from the most probable visited step:
\begin{equation}
n^* = \arg\max_{k \le n} p^{(k)}, \quad
A = A^{(n^*)}.
\end{equation}

\section{Experiments}

\subsection{Experiment Settings}

\begin{table}
  \caption{Main results on LIBERO across four task suites. 
$( L \otimes N)$ denotes the loop configuration.
Bold entries indicate our LoopVLA variants. 
Thrpt. denotes throughput (Hz), i.e., the average number of action predictions per second. 
For fairness, all methods are evaluated using eager attention without KV cache, except for methods marked with $\dagger$.
LoopVLA achieves consistent improvements in success rate while significantly reducing parameter count and improving throughput.Bold denotes the best result and underline denotes the second-best.}
  \label{tab:main_libero}
  \centering
  \footnotesize
  \begin{tabular}{ccccccccc}
    \toprule
    Model  & Spatial & Object & Goal & Long & Average &  Params $\downarrow$ &  FLOPs $\downarrow$ & Thrpt. (Hz) $\uparrow$ \\
    \midrule
    Diffusion Policy \cite{chi2025diffusion} & 78.3 & 92.5 & 68.3 & 50.0 & 72.4 & - & - & - \\
    OpenVLA \cite{kim2025openvla}$\dagger$ & 84.7 & 88.4 & 79.2 & 53.7 & 76.5 & 7.2B  & 6.58T & 3.26 \\
    $\pi_0$+FAST \cite{pertsch2025fast} & 96.4 & 96.8 & 88.6 & 60.2 & 85.5 & 3.5B & 606.68T & 0.05 \\
    GR00T-N1.5 \cite{nvidia2025gr00tn1} & 92.0 & 92.0 & 86.0 & 76.0 & 86.5 & 2.4B & 0.47T & 7.63 \\
    UniVLA \cite{deng2025graspvla}& 96.5 & 96.8 & 95.6 & \textbf{92.0} & 95.2 & 7.2B & 108.71T & 1.41 \\
    $\pi_0$ \cite{Black2025AV} & \underline{96.8} & 98.8 & 95.8 & 85.2 & 94.1 & 4.0B  & 1.79T & 3.70 \\
    $\pi_{0.5}$ \cite{Black2025AV} & 95.4 & 98.4 & \underline{97.2} & 89.6 & 95.1 & 4.1B  & 2.41T & 3.11 \\
    F1-VLA \cite{lv2025f1}$\dagger$ & \textbf{98.2} & 97.8 & 95.4 & \underline{91.3} & \underline{95.7} & 4.2B & 5.93T & 1.68 \\
    \cmidrule(){1-9}
    Qwen3FM & 94.0 & 92.3 & 91.3 & 65.7 & 85.8 & 2.3B & 0.53T & 0.97 \\
    \rowcolor{gray!12}
    LoopFM$\alpha$  $(3\otimes8)$ 
    & 91.0 
    & 99.0
    & 95.3 
    & 79.0 
    & 91.1 
    & \underline{1.3B} 
    & \underline{0.38T} 
    & 2.04 \\ 
    
    Qwen3OFT & 95.0 & 97.0 & 97.1 & 90.5 & 94.9 & 2.2B & 0.53T & 10.49 \\
    
    \rowcolor{gray!12}
    LoopOFT$_0$ $(3\otimes8)$  
    & 93.4 
    & 98.2 
    & 96.0 
    & 90.6 
    & 94.6 
    & \textbf{1.2B}  
    & \textbf{0.31T} 
    & \textbf{18.41} \\

    \rowcolor{gray!12}
    LoopOFT$\alpha$ $(3\otimes8)$  
    & 94.0 
    & \underline{99.6}
    & 96.8 
    & 91.0
    & 95.3 
    & \textbf{1.2B} 
    & 0.42T 
    & \underline{12.15} \\ 

    \rowcolor{gray!12}
    LoopOFT* $(3\otimes8)$ 
    & 95.0 
    & \textbf{100} 
    & \textbf{97.4} 
    & 91.0
    & \textbf{96.0} 
    & \textbf{1.2B} 
    & 0.53T 
    & 10.93 \\

    \bottomrule
  \end{tabular}
\vspace{-0.3cm}
\end{table}

\begin{table}
\caption{
Zero-shot results on LIBERO-Plus across multiple generalization factors. 
LoopOFT$\alpha$ maintains competitive performance with significantly fewer parameters, highlighting strong generalization and efficiency.
}
  \label{tab:libero-plus}
  \centering
  \footnotesize
  \begin{tabular}{cccccccccc}
    \toprule
    Model & Camera & Robot & Language & Light & Background & Noise & Layout & Total & Params$\downarrow$ \\
    \midrule
    OpenVLA \cite{kim2025openvla} & 0.8 & 3.5 & 23.0 & 8.1 & 34.8 & 15.2 & 28.5 & 15.6 & 7.2B\\
    UniVLA \cite{bu2025univla}& 1.8 & 46.2 & 69.6 & 69.0 & 81.0 & 21.2 & 31.9 & 45.8 & 7.2B \\
    $\pi_0$ \cite{Black2025AV} & 13.8 & 6.0 & 58.8 & 85.0 & 81.4 & \textbf{79.0} & 68.9 & 53.6 & 4.0B \\
    $\pi_0$+FAST \cite{pertsch2025fast}  & \underline{65.1} & 21.6 & 61.0 & 73.2 & 73.2 & \underline{74.4} & 68.8 & 61.6 & 3.5B \\
    $\pi_{0.5}$ \cite{pmlr-v305-black25a} & 53.0 & \underline{50.3} & 65.7 & 83.1 & 77.3 & 53.2 & \textbf{72.7} & 65.0 & 4.1B \\
    VLA-Adapter \cite{wang2026vla} & \textbf{76.6} & 36.4 & \textbf{73.8} & 71.0 & 70.2 & 37.4 & 57.2 & 60.4 & \textbf{1.2B}\\
    \cmidrule(lr){1-10}
    Qwen3OFT & 45.6 & \textbf{55.0} & \underline{73.5} & \underline{86.1} &  \textbf{90.2} & 69.3 & \underline{72.4} & \textbf{68.8} & \underline{2.2B} \\
    
    \rowcolor{gray!12}
    LoopOFT$_{\alpha}$ $(3\otimes8)$  
    & 58.3 & 41.7 & 66.7 & \textbf{88.9} 
    & \underline{88.3} & 61.4 & 71.8 & \underline{65.8} & \textbf{1.2B} \\
    \bottomrule
  \end{tabular}
\vspace{-0.3cm}
\end{table}

\paragraph{Simulation Benchmark Details.}
We evaluate on LIBERO \cite{liu2023libero}, VLA-Arena \cite{zhang2025vla}, and LIBERO-Plus \cite{fei2025libero}. 
LIBERO contains multiple task suites covering spatial reasoning, object interaction, and long-horizon manipulation, while VLA-Arena provides a standardized benchmark with diverse tasks and difficulty levels. 
To assess generalization, we additionally report zero-shot performance on LIBERO-Plus, which introduces unseen task compositions.

Following prior work, we report the average success rate over 50 trials per suite on LIBERO and the average success rate across all tasks on VLA-Arena. For LIBERO, models are jointly trained on the combined data from all four task suites and evaluated on each suite individually using the same checkpoint. For VLA-Arena, training uses only the VLA-Arena-L0-Small subset. For LIBERO-Plus, we report zero-shot success rates without additional finetuning.

\paragraph{Baselines.}
Our experiments are instantiated on a VLA architecture built upon Qwen3VL-2B-Instruction \cite{bai2025qwen3}, with a SigLIP-2 \cite{tschannen2025siglip} visual encoder and a 28-layer Transformer decoder, including a 3-layer DeepStack module. 
We freeze the first four layers together with the DeepStack module—referred to as the \emph{PALs}—to preserve low-level perceptual features, and apply looping only to the remaining 24 layers. This confines iterative refinement to higher-level representations where sufficiency is most relevant while maintaining stable early-stage features.

For action prediction, we adopt two main stream policy heads: an OFT head (OpenVLA-OFT \cite{kim2025finetuning}) and a flow-matching head (GR00T \cite{nvidia2025gr00tn1}), both kept in their original minimal forms. The primary experimental validation is conducted using the StarVLA framework \cite{community2026starvla}.

\paragraph{Implementation Details.}

For LoopVLA, we denote a configuration by $( L \otimes N)$, where $L$ is the number of retained Transformer layers and $N$ is the maximum number of loop iterations. All models are initialized from pretrained checkpoints, retaining only the first $k$ layers of the LLM decoder while discarding the rest.

For all simulation benchmarks, we perform full-parameter finetuning unless otherwise specified. Training is conducted on NVIDIA A100 GPUs, while inference is performed on a single NVIDIA RTX 4090 GPU.

For result annotation, $*$ indicates the optimal selection inference configuration, while subscript $\alpha$ denotes a trade-off inference strategy balancing performance and efficiency. A subscript with a number (e.g., $1$) represents inference with a fixed number of loops.

\subsection{Main Results on Simulation Environment}

\paragraph{Results on LIBERO.}
We evaluate LoopVLA on LIBERO under a unified setup with identical backbones, using a $(3 \otimes 8)$ configuration.
As shown in Table~\ref{tab:main_libero}, LoopVLA achieves strong performance while improving efficiency. LoopOFT* attains an average success rate of \textbf{96.0\%}, outperforming strong baselines such as $\pi_0$ and Qwen3OFT, despite using fewer parameters. 
Some baseline results are reported from or reproduced following VLA-Evaluation-Harness.~\cite{choi2026vlaeval}.

LoopVLA also improves performance on challenging settings, reaching \textbf{91.0\%} on Long-horizon tasks, while maintaining near-saturated results on Object tasks. The trade-off variant (LoopOFT$_\alpha$) achieves \textbf{95.3\%}, indicating robustness under reduced computation.
Compared to Qwen3OFT, our method significantly reduces model size while maintaining performance, and supports flexible performance--efficiency trade-offs across inference strategies. These gains are attributed to the iterative refinement design, which adaptively allocates computation across steps.

\begin{table}
\caption{
Results on VLA-Arena (L0 level) across four task categories. 
SmolVLA is trained on VLA-Arena-L0-Large, while the other methods are trained on VLA-Arena-L0-Small.
}
  \label{tab:vla_arena}
  \centering
  \begin{tabular}{lccccc}
    \toprule
    Method & Safety & Distractor & Extrapolation & Long Horizon & Average \\
    \midrule
    SmolVLA \cite{shukor2025smolvla} & 33.0 &  48.0 & \textbf{23.0} & 74.0 & 37.0 \\
    Qwen3OFT & 54.0  & \textbf{71.0}  & 13.3 & 59.0 & 47.0 \\
    \rowcolor{gray!12}
    LoopVLA  & \textbf{55.6} & 60.0 & 12.0 & \textbf{76.0} &  \textbf{48.7}  \\
    \bottomrule
  \end{tabular}
\vspace{-0.3cm}
\end{table}

\paragraph{Results on VLA-Arena.}
We evaluate LoopVLA on the VLA-Arena benchmark under the same experimental protocol as LIBERO. Results are reported in Table~\ref{tab:vla_arena}.

LoopVLA outperforms Qwen3OFT in average performance, with gains on Safety and Long Horizon tasks. Despite minor variations in other categories, the results indicate that LoopVLA generalizes effectively beyond LIBERO, while remaining efficient under the same training setup.

\paragraph{Generalization on LIBERO-Plus.}
We evaluate zero-shot robustness on LIBERO-Plus. As shown in Table~\ref{tab:libero-plus}, LoopVLA achieves \textbf{66.5\%} success, remaining competitive with larger models such as $\pi_{0.5}$ (65.0\%) while using fewer parameters, with a modest $\sim$3\% gap to the strongest baseline.

LoopVLA remains robust across variations, particularly under visual changes (e.g., lighting and background), though performance degrades on more challenging factors (e.g., robot variation). This trade-off reflects reduced model capacity and adaptive computation, suggesting that iterative refinement maintains stable generalization under distribution shifts with improved efficiency.

\subsection{Ablation Study}

\begin{table}[t]
\centering
\footnotesize

\begin{minipage}[t]{0.48\linewidth}
\centering
\caption{Performance impact of loop configurations (Stage 1 training only).}
\label{tab:loop_cfg}
\vspace{5pt}
\begin{tabular}{ccccccc}
\toprule
Config & S & O & G & L & Avg & Params$\downarrow$ \\
\midrule
$(8\!\otimes\!3)$  & 95.0 & \textbf{100} & 95.3 & 80.7 & 92.8 &  1.48B \\
$(6\!\otimes\!4)$  & 96.0 & 98.7 & 95.7 & 85.0 & 93.9 &  1.37B \\
$(4\!\otimes\!6)$  & \textbf{96.4} & 96.2 & \textbf{97.0} & \textbf{90.5} & \textbf{95.0} & 1.27B \\
$(3\!\otimes\!8)$  & 94.7 & 98.6 & 96.6 & 90.2 & \textbf{95.0} & 1.22B \\
$(2\!\otimes\!12)$ & 91.2 & 98.6 & 95.0 & 83.4 & 92.1 & \textbf{1.17B} \\
\bottomrule
\end{tabular}
\end{minipage}
\hfill
\begin{minipage}[t]{0.48\linewidth}
\centering
\caption{Impact of inference layer settings (both Stage 1 and 2 training).}
\label{tab:infer_cfg}
\begin{tabular}{cccccc}
\toprule
Setting & S & O & G & L & Avg \\
\midrule
$(3\!\otimes\!8)_1$  & 91.0 & 98.7 & 95.4 & 90.0 & 93.8 \\
$(3\!\otimes\!8)_3$  & 93.2 & 99.4 & 96.0 & 90.6 & 94.9 \\
$(3\!\otimes\!8)_5$  & 94.6 & 96.7 & 96.2 & \textbf{91.0} & 94.7 \\
$(3\!\otimes\!8)_6$  & 93.4 & 99.4 & 95.6 & 90.6 & 94.8 \\
$(3\!\otimes\!8)_8$  & 92.0 & 99.0 & 94.6 & \textbf{91.0} & 94.2 \\
$(3\!\otimes\!8)^*$  & \textbf{95.0} & \textbf{100} & \textbf{97.4} & \textbf{91.0} & \textbf{96.0} \\
\bottomrule
\end{tabular}

\end{minipage}
\vspace{-0.3
cm}
\end{table}

\paragraph{Effect of Loop Configurations.}
We study how to allocate a fixed computation budget by varying loop configurations $(L \otimes N)$ under $L \times N = 24$. As shown in Table~\ref{tab:loop_cfg}, performance follows a non-monotonic trend. Here, S, O, G, L, and Avg denote the Spatial, Object, Goal, Long, and average task suites, respectively. Balanced configurations (e.g., $(4 \otimes 6)$ and $(3 \otimes 8)$) achieve the best results, while more extreme settings degrade performance. This suggests that neither increasing depth nor iteration alone is sufficient; instead, effective performance arises from a proper balance between the two. We adopt $(3 \otimes 8)$ as the default configuration.

\paragraph{Inference Layer Selection.}
Under a fixed $(3 \otimes 8)$ configuration, we evaluate different inference layer selections (Table~\ref{tab:infer_cfg}). Performance again exhibits a non-monotonic pattern: both shallow and overly deep selections underperform, while intermediate layers yield better results. Notably, adaptive selection ($*$) consistently outperforms fixed strategies, indicating that the optimal computation depth is input-dependent. The variation of optimal depths across tasks further supports the effectiveness of Stage~2 training. We further visualize the distribution of maximum sufficiency across different LIBERO task suites in Appendix Sec.~\ref{sec:distribution}, highlighting task-dependent variation in refinement behavior.

\begin{table}
\caption{
Ablation of the sufficiency head on LIBERO, comparing a standard MLP against the proposed sufficiency-aware design.
}
  \label{tab:suf_ablation}
  \centering
  \begin{tabular}{lccccc}
    \toprule
    Method & Spatial & Object & Goal & Long & Average \\
    \midrule
    Direct MLP & 94.0 & 99.2  & 97.2 & 87.0 & 94.4 \\   
    \rowcolor{gray!12}
    Sufficiency Head & \textbf{95.0} & \textbf{100.0} & \textbf{97.4} & \textbf{91.0} & \textbf{96.0} \\
    \bottomrule
  \end{tabular}
\vspace{-0.3cm}
\end{table}

\paragraph{Effect of the Sufficiency Head.}
We replace the proposed sufficiency head with a standard MLP. 
As shown in Table~\ref{tab:suf_ablation}, this leads to a consistent performance drop, especially on long-horizon tasks. 
This suggests that a simple MLP is insufficient to capture the notion of representational sufficiency. 
In contrast, the proposed design explicitly models interactions between sufficiency and action tokens, enabling more accurate sufficiency estimation and leading to improved overall performance.

\section{Conclusion}

In this work, we revisit VLA policy learning from the perspective of representational sufficiency. Rather than relying on fixed-depth representations or heuristic stopping criteria, we formulate action prediction as an iterative refinement process in which representation quality and decision sufficiency are jointly modeled.

Based on this formulation, we propose LoopVLA, a recurrent VLA architecture with a shared looped Transformer, a dual-head design, and a distribution alignment objective for sufficiency learning. By decoupling representation refinement from absolute layer depth, LoopVLA achieves a substantially improved efficiency--performance trade-off, reducing model parameters while maintaining or improving policy performance across multiple embodied benchmarks.

We hope this work highlights the importance of adaptive representation refinement in embodied policy learning and motivates future research on scalable, content-aware, and computation-efficient VLA architectures.



\newpage
\bibliographystyle{unsrt}
\bibliography{references}


\newpage
\appendix
\section{Appendix}

\subsection{LIBERO-Plus Benchmark}

We evaluate the generalization capability of LoopVLA on LIBERO-Plus, a benchmark designed for in-depth robustness analysis of Vision-Language-Action models under controlled distribution shifts.

\begin{figure}[!htbp]
    \centering
    \includegraphics[width=0.9\linewidth]{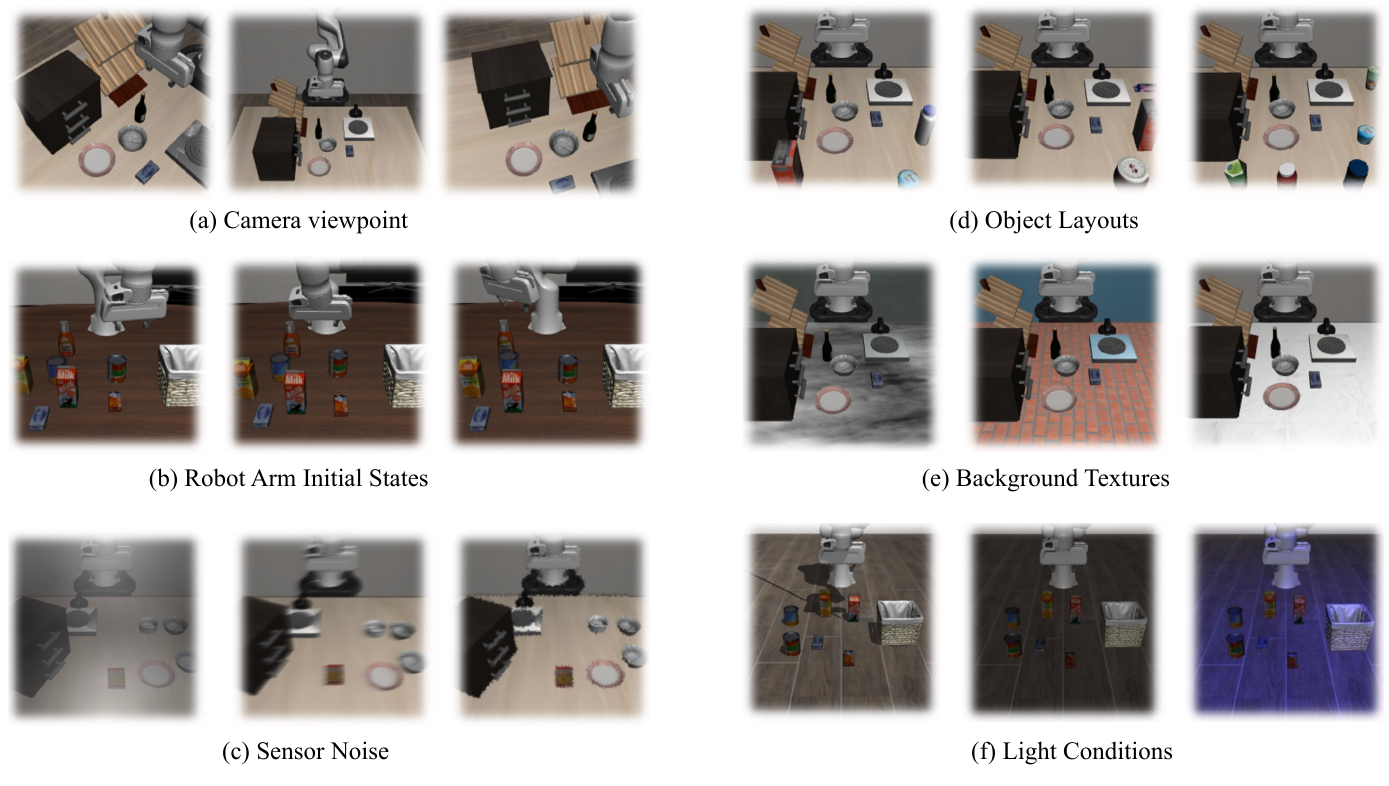}
    \caption{
Examples of distribution shifts in LIBERO-Plus across different generalization dimensions: 
(a) Camera viewpoints, varying the observation perspective; 
(b) Robot arm initial states, changing the robot's starting configuration; 
(c) Sensor noise, introducing visual perturbations and corruption; 
(d) Object layouts, altering spatial arrangements of objects; 
(e) Background textures, modifying scene appearance; 
(f) Light conditions, varying illumination settings. 
These controlled perturbations evaluate robustness from low-level visual variations to higher-level spatial and embodiment-related changes.
}
    \label{fig:libero-plus}
\end{figure}

LIBERO-Plus performs a systematic evaluation by introducing perturbations along seven dimensions: object layout, camera viewpoints, robot initial states, language instructions, lighting conditions, background textures, and sensor noise. These factors cover a wide spectrum of variations, ranging from low-level visual changes (e.g., lighting, background, noise) to higher-level variations in scene composition, viewpoint, and instruction semantics. Representative examples are shown in Figure~\ref{fig:libero-plus}.

All evaluations are conducted in a zero-shot setting without any additional fine-tuning, following the official evaluation protocol. Performance is measured by the average task success rate across tasks within each perturbation dimension, as well as the overall average across all dimensions. This setup enables a comprehensive assessment of model robustness and reliability under realistic variations beyond the standard LIBERO benchmark.

\subsection{VLA-Arena Benchmark}

We further evaluate LoopVLA on VLA-Arena, an open-source benchmark for systematic evaluation of Vision-Language-Action (VLA) models. VLA-Arena provides a full evaluation pipeline, including scene modeling, demonstration collection, model training, and standardized evaluation.

\begin{figure}
    \centering
    \includegraphics[width=0.9\linewidth]{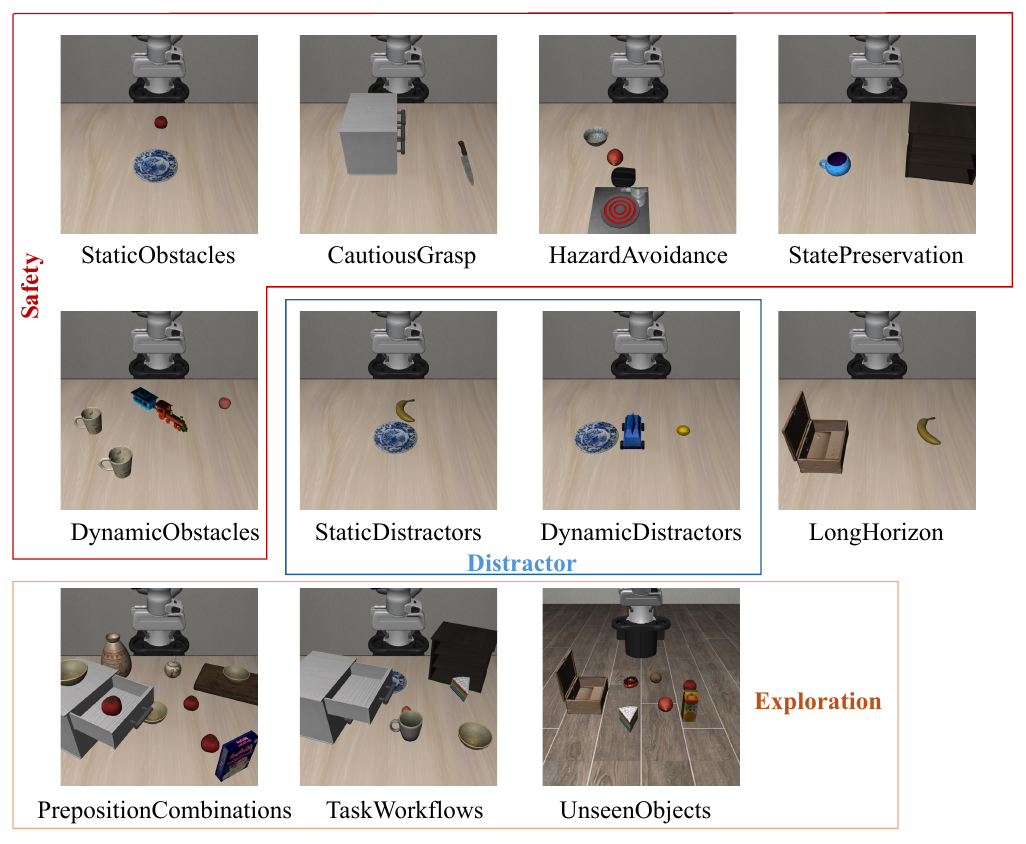}
    \caption{
    Representative L0 task suites in VLA-Arena across four domains: safety (top), distractors (middle), extrapolation (bottom), and long-horizon tasks (right). These tasks evaluate reliability, robustness, generalization, and long-horizon reasoning.
    }
    \label{fig:vla-arena}
\end{figure}

The benchmark consists of 170 tasks organized into 11 specialized task suites, with hierarchical difficulty levels ranging from L0 to L2. It is designed to assess model performance across four key domains: (i) safety, requiring reliable and safe interaction with the environment; (ii) distractors, evaluating robustness to environmental variability; (iii) extrapolation, measuring generalization to novel task compositions; and (iv) long-horizon reasoning, testing the ability to execute extended action sequences. Example tasks are illustrated in Figure~\ref{fig:vla-arena}.

In our experiments, we focus on the L0 setting and use the VLA-Arena-L0-Small subset for both training and evaluation. All methods are evaluated under a unified protocol with consistent inference settings and success metrics. Performance is reported using the average task success rate across all task suites following the official benchmark setup.

\subsection{Distribution of Optimal Loop Indices}
\label{sec:distribution}

\begin{figure}
    \centering
    \includegraphics[width=0.9\linewidth]{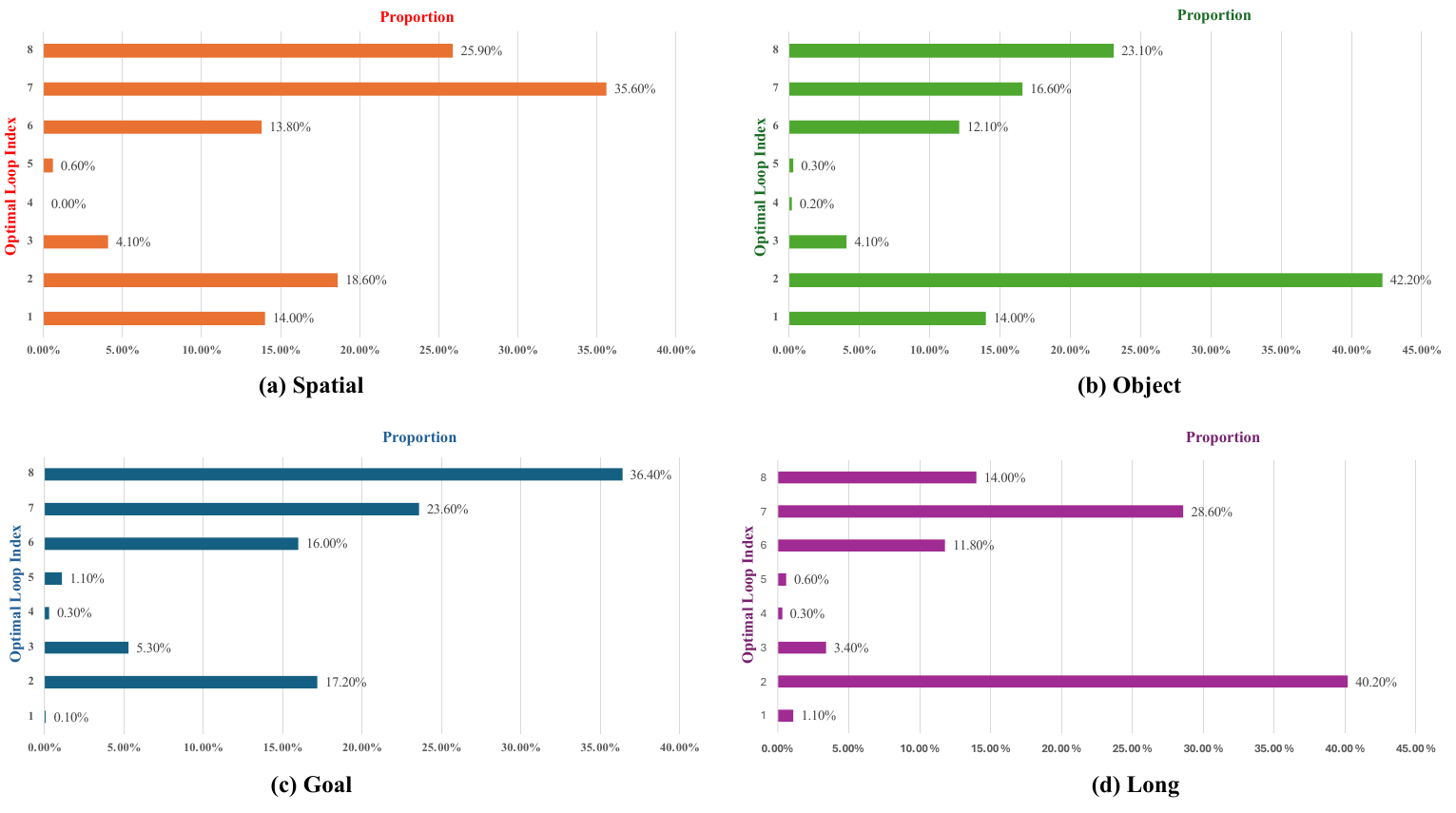}
    \caption{
    Distribution of selected loop indices $n^*$ across different LIBERO task suites, computed from sampled evaluation episodes. 
    Different task suites exhibit distinct distribution patterns over loop indices, reflecting varying computational demands. 
    }
    \label{fig:loop_index_dist}
\end{figure}

We analyze the distribution of selected loop indices $n^*$ across different LIBERO task suites, as shown in Fig.~\ref{fig:loop_index_dist}. 
The statistics are obtained from uniformly randomly sampled episodes within each suite, rather than exhaustively evaluating the full benchmark.

A clear distinction emerges across task types. Simpler suites such as \textit{Object} show a strong concentration on early loop indices (e.g., $n=2$), indicating that minimal refinement is sufficient. 
In contrast, more complex suites such as \textit{Spatial} and \textit{Goal} exhibit distributions skewed toward larger loop indices (e.g., $n=7,8$), suggesting a greater need for iterative refinement. 
The \textit{Long} suite presents a mixed pattern, with both early stopping and deeper refinement occurring frequently, reflecting the variability of long-horizon tasks.
This also explains the low frequency of intermediate loop indices (e.g., $n=3,4$), as the model tends to either terminate early when sufficient information is obtained or continue refinement toward later steps for more complex decisions, resulting in a bimodal preference.

Overall, these results support that LoopVLA dynamically adjusts its inference depth according to task difficulty, rather than relying on a fixed number of refinement steps.

\subsection{Training Hyperparameters}

We follow the training and inference setup described in Sec.~\ref{sec:train_detail} and Sec.~\ref{sec:infer_detail}. 
All hyperparameters are summarized in Table~\ref{tab:hyperparams}.

In brief, models are trained using AdamW with a cosine learning rate schedule and a global batch size of 64. Training of stage 2 is typically updated for 10K steps, or equivalently one epoch over the dataset.
Entropy and diversity regularization are applied during early training to stabilize the halting behavior. 

All experiments are implemented in PyTorch and trained on 8 NVIDIA A100 (40GB) GPUs. 
For fair comparison, all methods use the same backbone architectures and data protocols as their corresponding baselines.

\begin{table}[t]
\caption{
Training and inference hyperparameters.
}
\label{tab:hyperparams}
\centering
\begin{tabular}{lc}
\toprule
\textbf{Hyperparameter} & \textbf{Value} \\
\midrule
\multicolumn{2}{l}{\textit{Training}} \\
Optimizer & AdamW \\
LR Scheduler & Cosine \\
Global Batch Size & 64 \\
Action Chunk Size & 8 \\
Action Tokens & 8 \\
Sufficiency Tokens & 3 \\

Entropy Reg. $\lambda_1$ & 0.001 \\
Diversity Reg. $\lambda_2$ & 0.01 \\
Warmup Steps & 1K \\
Temperature $\tau$ & 0.5 \\
\midrule
\multicolumn{2}{l}{\textit{Inference}} \\
Threshold $\theta$  & 0.68 \\
\bottomrule
\end{tabular}
\end{table}

\subsection{Limitations}
Although LoopVLA improves the efficiency-performance trade-off of VLA models, several limitations remain. 
First, while the proposed recurrent refinement mechanism demonstrates promising robustness and generalization ability, handling more severe distribution shifts and embodiment variations may require additional scaling in model capacity and training data. 

Second, due to computational resource constraints, the current study does not extensively explore scaling behaviors across larger model sizes and data scales. A more systematic investigation of scaling properties may further reveal the potential and limitations of the proposed framework.

Finally, the current study primarily focuses on simulation benchmarks, and future work will extend the framework to real-world robotic platforms and broader embodied manipulation settings.



\end{document}